%% file: main.tex
\documentclass{article} 
\usepackage{iclr2024_conference,times}
\newcommand{\Modelfull}{Generative Neuro-symbolic Visual Reasoning Model\xspace}
\newcommand{\Model}{GENOME\xspace}

\newcommand{\ModuleA}{module initialization\xspace}
\newcommand{\ModuleB}{module generation\xspace}
\newcommand{\ModuleC}{module execution\xspace} 

\newcommand{\ModuleAA}{Module Initialization\xspace}
\newcommand{\ModuleBB}{Module Generation\xspace}
\newcommand{\ModuleCC}{Module Execution\xspace} 

\definecolor{MyDarkRed}{rgb}{0.8,0.02,0.02}

\makeatletter
\DeclareRobustCommand\onedot{\futurelet\@let@token\@onedot}
\def\@onedot{\ifx\@let@token.\else.\null\fi\xspace}

\def\eg{e.g\onedot} 
\def\ie{i.e\onedot} 
 
\def\etc{etc\onedot}

\makeatother

\usepackage[colorlinks,citecolor=gray,linkcolor=red]{hyperref} \usepackage{graphicx} 
\usepackage{url}
\usepackage{wrapfig}
\definecolor{Gray}{gray}{0.9}
\usepackage{xspace}
\usepackage{booktabs}
\usepackage{multirow}
\usepackage[utf8]{inputenc}
\usepackage{listings}
\usepackage[frozencache,cachedir=.]{minted}

\definecolor{c1}{RGB}{128, 237, 18}
\definecolor{c2}{RGB}{165, 214, 4}
\definecolor{c3}{RGB}{199, 182, 1}
\definecolor{c4}{RGB}{227, 146, 9}
\definecolor{c5}{RGB}{246, 108, 28}
\definecolor{c6}{RGB}{246, 87, 66}

\title{\textcolor{c1}{G}\textcolor{c2}{E}\textcolor{c3}{N}\textcolor{c4}{O}\textcolor{c5}{M}\textcolor{c6}{E}: \textcolor{c1}{G}enerativ\textcolor{c2}{E} \textcolor{c3}{N}euro-Symb\textcolor{c4}{O}lic Visual Reasoning by Growing and Reusing \textcolor{c5}{M}odul\textcolor{c6}{E}s}
\author{%
      Zhenfang Chen \thanks{indicates equal contributions} \\
      MIT-IBM Watson AI Lab\\
   \And
      Rui Sun$^{*}$ \\
      Columbia University \\
     \And
      Wenjun Liu$^{*}$ \\
      Tsinghua University\\
     \And
     Yining Hong \\
     University of California, Los Angeles \\
    \And
      Chuang Gan \\
      MIT-IBM Watson AI Lab and UMass Amherst \\
 }

\iclrfinalcopy 
\begin{document}
\maketitle

\vspace{-0.5em}
\begin{abstract}
Recent works have shown that Large Language Models (LLMs) could empower traditional neuro-symbolic models via programming capabilities to translate language into module descriptions, thus achieving strong visual reasoning results while maintaining the model's transparency and efficiency. However, these models usually exhaustively generate the entire code snippet given each new instance of a task, which is extremely ineffective. On the contrary, human beings gradually acquire knowledge that can be reused and grow into more profound skills for fast generalization to new tasks since we are an infant. 
Inspired by this, we propose generative neuro-symbolic visual reasoning by growing and reusing modules. Specifically, our model consists of three unique stages, \ModuleA, \ModuleB, and \ModuleC. 
First, given a vision-language task, we adopt LLMs to examine whether we could reuse and grow over established modules to handle this new task. If not, we initialize a new module needed by the task and specify the inputs and outputs of this new module.
After that, the new module is created by querying LLMs to generate corresponding code snippets that match the requirements. In order to get a better sense of the new module's ability, we treat few-shot training examples as test cases to see if our new module could pass these cases. If yes, the new module is added to the module library for future reuse. Finally, we evaluate the performance of our model on the testing set by executing the parsed programs with the newly made visual modules to get the results. 
We find the proposed model possesses several advantages.
First, it performs competitively on standard tasks like visual question answering and referring expression comprehension;
Second, the modules learned from one task can be seamlessly transferred to new tasks;
Last but not least, it is able to adapt to new visual reasoning tasks by observing a few training examples and reusing modules\footnote{Project page: \url{https://vis-www.cs.umass.edu/genome}}.
\end{abstract}

\vspace{-0.5em}
\section{Introduction}
\vspace{-0.5em}
Neuro-symbolic visual reasoning models~\citep{andreas2016neural,mao2019neuro} refer to the algorithm family that combines deep neural networks~\citep{lecun1998gradient,hochreiter1997long} for learning correlations among the training data and symbolic methods~\citep{yi2018neural,andreas2016learning} to perform explicit and transparent multi-step reasoning.
In contrast to pure neural network-based models~\citep{hudson2018compositional,li2023blip}, neuro-symbolic approaches achieve strong performance in visual reasoning tasks, simultaneously offering superior model transparency and data efficiency.

Nevertheless, such models suffer from several inherent limitations. Firstly, their language parsers~\citep{yi2018neural,andreas2016neural}, employed for the conversion from natural language into symbolic programs, typically demand extensive domain-specific language-program pairs to train on, and struggle to generalize effectively to unconstrained natural language instructions. Additionally, these models necessitate a custom design for every module, rendering the process labor-intensive and lacking scalability.

Recent advancements in large language models (LLMs) ~\citep{brown2020language,ouyang2022training} have ushered in a new era with its remarkable performances across various applications,
including chatbots \citep{shuster2022blenderbot}, virtual assistants ~\citep{dong2023towards}, and programming assistants ~\citep{chen2021evaluating}. 
Riding this unprecedented wave, researchers reformulate the old wisdom by incorporating LLMs into neuro-symbolic reasoning, bypassing the inflexibility and ineffectiveness of domain-specific language-to-program parsers. Specifically, VisProg~\citep{Gupta2022VisProg} pre-defines a set of visual modules, and uses LLMs to transform language instructions into symbolic programs consisting of the pre-defined visual modules. 
Taking a step forward, ViperGPT ~\citep{suris2023vipergpt} releases the burden on manually-defined visual modules by introducing a code generator that could produce a code snippet based on each input instance of a new task. 

Promising as these LLM-based neuro-symbolic models can be, they inevitably bear several weaknesses compared to the learning and reasoning processes of human beings.
First, both VisProg and ViperGPT exhaustively produce one code snippet for each new instance of a task, which is extremely ineffective. This is in stark contrast with the human learning process: from an early age, we organically accumulate knowledge from particular experiences. Such knowledge acquired from specific cases could be reused and reconfigured, enabling us to quickly adapt to new tasks and new demands ~\citep{Harlow1949TheFO, Mitchell1986ExplanationBasedGA, lake2016building, ellis2023dreamcoder}. The knowledge blocks grow progressively over time, gradually into a library with extraordinary richness and flexibility for fast generalization to any unseen task - the knowledge library that these models fall short of. Second, both models do not verify and examine the codes they generate. It seems that when the models generate a bad code snippet that cannot solve the input case, they just ``let it go" without taking another stab for larger chance towards success. And of course, when they encounter similar cases again, they keep ``stepping on the same rake". Human beings, on the other hand, would verify and examine the acquired knowledge by proposing a set of test scenarios before storing them in the library \citep{Brul1989KnowledgeA}. It's crucial that a neuro-symbolic reasoning model is equipped with the same abilities to verify the codes it produces, stores them in a library if satisfactory, and makes another attempt when the codes fail.

\begin{figure}[t]
    \centering
    \includegraphics[width=.99\linewidth]{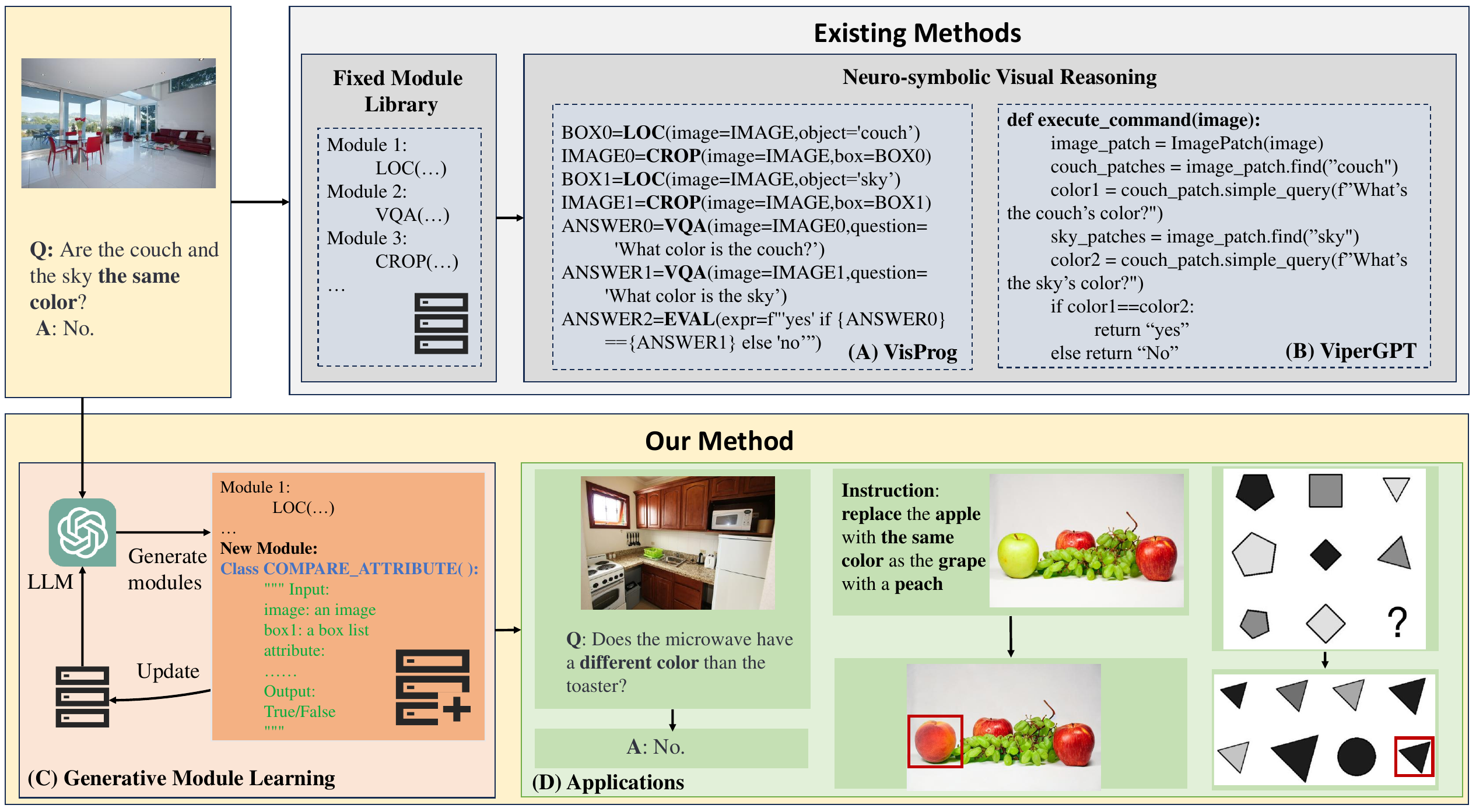}
    \vspace{-0.5em}
    \caption{The motivation of the \Model. Compared with VisProg and ViperGPT which exhaustively generate a code snippet for each input case, our \Model is able to generate new modules and reuse old modules to handle the query. The module generated by \Model can be used to handle other instances of the task for better performance. Second, the generated module can be transferred to different tasks like image editing. Finally, it can learn to handle new tasks like Raven~\citep{burke1985raven,zhang2019raven} by learning modules from only a few training samples. The edited region and the correct answer for the Raven task are labeled with red boxes for better visualization.}
    \label{fig:teaser}
\end{figure}

To this end, we introduce a novel  \Modelfull (\Model), proficient in assimilating new neural modules from a limited set of training examples.
This model excels in handling standard visual reasoning tasks such as visual question answering. Additionally, it demonstrates outstanding module transfer capabilities for tasks like image editing, and exhibits an exceptional ability to generalize to new reasoning tasks with limited training examples.
As illustrated in Figure~\ref{fig:model}, \Model comprises three stages: 1) \ModuleA, 2) \ModuleB, and 3) \ModuleC.
From the initial training set examples, an LLM discerns the necessity for a new module to tackle the task and, if required, produces the respective input and output.
In the second stage, LLMs implement and refine the new module, ensuring seamless integration with existing modules and resulting in accurate responses to the training queries.
During testing, the LLM first converts language instructions into executable high-level programs like \textit{COMPARE\_ATTRIBUTE(IMAGE,BOX0,BOX1,ATTR)} for comparing attributes of different bounding boxes. The program will be run with the new module sets, producing the desired outputs.

We assessed the performance of \Model across six visual reasoning tasks, spanning from visual question answering~\citep{hudson2019gqa} to Raven's Progressive Matrices~\citep{zhang2019raven}. The experimental findings reveal that \Model delivers competitive results on standard benchmarks, ensuring both transparency and interoperability. Notably, modules honed on these standard tasks can be adeptly adapted to diverse domains, including image editing and knowledge tagging~\citep{Gupta2022VisProg}. Additionally, with minimal training examples, \Model demonstrates the capability to manage new visual reasoning tasks~\citep{burke1985raven,jiang2023mewl} by repurposing modules. Our code and data will be publicly available.

\section{Related Work}
\textbf{Visual Reasoning.}
Our work aims to handle visual reasoning tasks, which require a model to draw new inferences based on the acquired visual cues in images or videos~\citep{hudson2019gqa,kazemzadeh2014referitgame,Goyal_2017_CVPR,zhang2019raven,jiang2023mewl}. Typical tasks for visual reasoning include visible question answering~\citep{Goyal_2017_CVPR,hudson2019gqa}, visual grounding~\citep{kazemzadeh2014referitgame,yu2016modeling,chen2020cops} and Raven’s Progressive Matrices~\citep{burke1985raven,zhang2019raven}. Various models~\citep{hudson2018compositional,yu2018mattnet,zhang2021abstract,ding2023embodied} have been developed to handle these tasks but most of them are ad-hoc and carefully designed for a specific task, leaving it an open research question on how to build a general model that can handle different kinds of visual reasoning problems by only showing a few examples.

\textbf{Neuro-symbolic Visual Reasoning.}
Our work is also closely related to neuro-symbolic visual reasoning models~\citep{andreas2016neural,Mao2019NeuroSymbolic,chen2021grounding,chen2022comphy}, where the models decompose the query of the visual reasoning tasks into a series of reasoning steps and represent each reasoning step with a neural module (\ie, a code snippet for achieving specific functions like localizing objects and recognizing object categories). While these models have better model interoperability and data efficiency than previous connectionist models~\citep{hudson2018compositional,anderson2018bottom}, they often show their limitations in representing natural language instructions in the wild with the limited pre-defined reasoning steps~\citep{yang2020object,chen2019meta}. Moreover, they need to manually define and implement each neural module one by one, making it hard to scale up and handle multiple tasks within a single model.

\textbf{Foundation Models for Reasoning.}
Recently, large language models (LLMs)~\citep{brown2020language,ouyang2022training} have been widely used in language understanding~\citep{hendrycks2020measuring} and reasoning~\citep{cobbe2021gsm8k,amini2019mathqa}.~\citet{schick2023toolformer} develop the toolformer to show that LLMs can use external tools to better handle language tasks.~\citet{cai2023large} shows that LLMs can make simple tools for natural language tasks by writing code snippets.
LLMs have also been used in vision-language tasks. Most of these works~\citep{li2023blip,alayrac2022flamingo} connect LLMs with additional vision encoders and fine-tune them with massive vision-language pairs. As evaluated by ~\citet{xu2023lvlm}, while these models show great performance on in-domain tasks, they also perform poorly on tasks out of the training domains. They are also extremely computation-expensive. For example, it takes 15 days to use 1536 TPUv4 for training Flamingo~\citep{alayrac2022flamingo}. 

\textbf{Visual Programming by LLMs.}
Another line of research has been combining vision models~\citep{li2021grounded,kirillov2023segany,Radford2021LearningTV} with LLMs in an off-shelf manner. Early models~\citep{yang2022empirical,chen2023see} transformed images into captions and append the captions into the LLMs' prompt to handle vision-language tasks. While the method is simple, they also perform inferior and lack transparency. Recently, VisPROG~\citep{Gupta2022VisProg} uses LLMs to transform language instructions into pre-defined modular operations for step-by-step reasoning. However, it still requires manually implementing each module one by one. Later, ViperGPT~\citep{suris2023vipergpt} shows that the LLMs can be used to write a code snippet for each query instance independently to handle vision tasks. However, the code it writes has not been examined and tested by any training examples and there is no guarantee about the performance and code safety. Instead, we propose \Model that ask LLMs to create new neural models (\ie general code snippets to achieve specific functions) and handle the given tasks through only a few training examples. Such newly generated modules can cooperate with each other and be reused for other tasks for better performance.

\section{Method}
\begin{figure}[t]
    \centering
    \includegraphics[width=\linewidth]{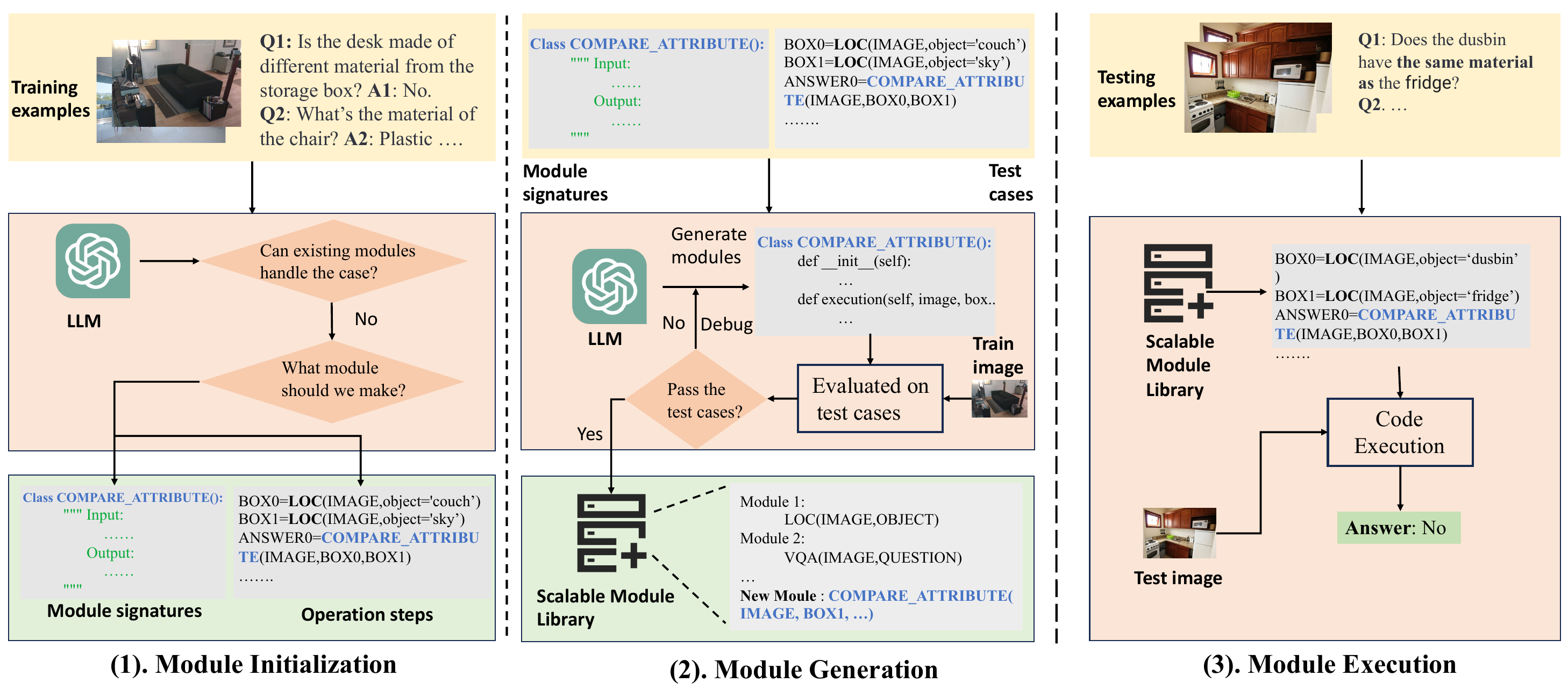}
    \caption{
The framework of our \Model, which contains three stages, \ModuleA, \ModuleB, and \ModuleC. In stage 1, we feed the questions and the signature of the existing modules to the LLM and ask it to identify whether we can handle the query within the operations of the existing modules. If not, we ask the LLM to generate the signature of the new module (\ie the input and output) and predict the reasoning steps to handle the query task. In stage 2, we feed the module signature and the testing cases to the LLM and ask the LLM to implement the module and test its pass rate on the training examples. We only accept the modules that successfully handle the query. In stage 3, we first use the LLM to parse the query into symbolic operations and then execute these operations on the test images with the help of the scalable module library. We take VQA as an example and such a framework can also be expanded to other tasks like referring expression comprehension and Raven.
     }
    \label{fig:model}
\end{figure}
\subsection{Overall}
In this section, we present a novel framework named as \Modelfull (\Model) for the acquisition of neural modules and solutions of visual reasoning tasks with only a limited set of training examples.
\Model comprises several pre-defined operators that serve as the initial building blocks. Each neural operator corresponds to a neural module and is implemented using a Python code snippet, thereby enabling specific functions such as object localization within an image.
Nevertheless, it is not possible to pre-define all the necessary neural modules prior to addressing the visual reasoning tasks. Consequently, there arises a need to generate new modules based on a limited number of visual reasoning task examples. 

Figure~\ref{fig:model} illustrates that \Model consists of three distinct stages: 1) \ModuleA, 2) \ModuleB, and 3) \ModuleC.
In the \ModuleA stage, when provided with a small set of training examples from the visual reasoning dataset, the primary objective is to determine whether the existing neural modules are sufficient to address the query examples.
If the existing modules are inadequate for the query instance, \Model will identify the requirements for creating new modules, including defining the modules' input and output specifications.
During the \ModuleB stage, \Model leverages the LLM to implement the neural module based on the provided training examples and the specified input and output format (\ie, function signature) and add the module to the module library only when it passes the test cases.
Once the new module is successfully implemented, the \ModuleC orchestrates the transformation of input queries into a sequence of neural operations. Subsequently, these operations are applied to the neural modules to obtain the correct output. All these three stages are powered by robust code generation capabilities and the in-context learning technique. Prompts for each stage can be found at Figure~\ref{prompt:stage1}-\ref{prompt:stage3}

\subsection{Model Details}
Utilizing a limited number of examples from the training set of visual reasoning tasks, we employ the \Model framework, comprising three distinctive stages: \ModuleA, \ModuleB, and \ModuleC. 

\paragraph{\ModuleAA.} The initial phase within our \Model framework is \ModuleA, dedicated to determining the set of new modules required to address the visual reasoning task. As depicted in Figure~\ref{fig:model}-1, we employ an LLM to assess the feasibility of handling these training instances using existing neural modules. Should this not be achievable, we task the LLM with specifying the necessary new modules (\eg \textit{COMPARE\_ATTRIBUTE} in Figure~\ref{fig:model}) for an accurate response to the query. The outcome of this stage comprises function signatures detailing the input and output formats for these new modules. Furthermore, it facilitates the transformation of the input query into a sequence of reasoning steps, which function as test cases to validate the correctness of the generated program within \ModuleB. The prompt for the LLM in this stage is shown in Figure~\ref{prompt:stage1}. 

\paragraph{\ModuleBB.}
The second phase of our \Model framework is \ModuleB, which focuses on implementing the correct new modules proposed during the \ModuleA stage. Specifically, after receiving the signature of a new module, we incorporate corresponding test cases into the prompt and employ learning-in-context techniques to generate multiple program candidates. These program candidates are subsequently executed using the provided training examples. If a program encounters errors during execution, we incorporate the error information into the LLM's prompt and instruct it to rectify these issues. We only accept program candidates that achieve a pass rate surpassing a predefined threshold ($\eta$). This procedure bears resemblance to the code translation of LLMs discussed in~\citep{chen2023teaching}, but we extend it to accommodate more intricate multi-modal input types and instructions from natural language and raw images. The inclusion of module generation in the context of visual reasoning tasks offers two principal advantages. Firstly, it upholds the transparency and interpretability of neuro-symbolic models while preserving competitive performance. Secondly, it exhibits generative capabilities and scalability as our \Model can autonomously generate new modules tailored to specific tasks.

\paragraph{\ModuleCC.}
Given the integration of newly-generated modules with existing neural modules tailored for visual reasoning, the \Model framework initiates query parsing from the testing dataset, transforming them into executable operations through in-context learning. An illustrative prompt for this stage is depicted in Figure~\ref{prompt:stage3}. Notably, although various visual reasoning tasks may possess distinct inputs and outputs, they can re-purpose these intermediary modules designed for other tasks to enhance overall performance. This feature represents a unique capability for code generation at the module level, an aspect hitherto unexplored by prior methods\citep{suris2023vipergpt,Gupta2022VisProg}.

\section{Experiments}
In this section, we present a comprehensive series of experiments to evaluate the performance of our models. Initially, we demonstrate our models' effectiveness in learning neural modules on two established benchmarks: GQA~\citep{hudson2019gqa}, focusing on compositional visual question answering, and RefCOCO~\citep{kazemzadeh2014referitgame}, which assesses referring expression comprehension. Subsequently, we illustrate how the modules acquired from these two datasets can be successfully applied to novel tasks such as image editing and knowledge tagging. Moreover, we highlight the adaptability of our framework to address novel visual reasoning tasks (Raven~\citep{zhang2019raven} and MEWL~\citep{jiang2023mewl}), even with limited training examples. Before delving into these experiments, we provide an overview of the experimental settings.

\noindent{\textbf{Experimental Details.}}
The success of our \Model relies on a set of pre-defined modules and APIs as the starting point. We utilize handcrafted modules from VisProg~\citep{Gupta2022VisProg} as our initial components. Additionally, we incorporate several new APIs from ViperGPT to enhance module creation. We also include some new APIs from ViperGPT~\citep{suris2023vipergpt} for making new modules. In Section~\ref{subsec:abs}, we also include results parsed by the open-source LLM from WLM~\citep{xu2023wizardlm} to investigate the influence of different LLM models.
A comprehensive list of the pretrained modules employed in our approach can be found in Section~\ref{imp:detail} of the Appendix.
We extract training examples to acquire new modules. More precisely, we extracted 300 examples from GQA, 100 from RefCOCO, 10 from Raven, and 10 from MEWL.

\noindent{\textbf{Datasets and Evaluation Metric.}}
We show experiments of our \Model on standard vision-language benchmarks,
GQA~\citep{hudson2019gqa} and RefCOCO~\cite{kazemzadeh2014referitgame}. GQA is a popular compositional visual reasoning dataset with synthesis multi-hop questions, making it suitable for multi-step reasoning. RefCOCO is a typical visual grounding dataset, evaluating models' ability to localize objects and understand fine-grained spatial and semantic relationships. Following ViperGPT, we evaluate GQA on test-dev split and RefCOCO on the testA split.
Then, we show \Model's abilities on other the transferred tasks, image editing, and knowledge tagging and compare it with VisProg. Since the image editing and knowledge tagging datasets from VisProg are not publicly available, we built two new datasets for evaluation. The new editing dataset contains 50 images and instruction pairs. The new knowledge tagging dataset contains 50 images with 50 referring expressions. We provide more details about the dataset in Appendix~\ref{dataset}. The datasets will be released for research purposes. Finally, we show that \Model can learn to handle new visual reasoning tasks like Raven~\citep{zhang2019raven} and MEWL~\citep{jiang2023mewl} by observing a few training examples and module learning. Raven is a task for relational and analogical visual reasoning of image sets and has been widely used for non-verbal intelligence tests. MEWL is a recent benchmark proposed to assess how machines learn word meaning in grounded visual scenes.
Examples of these tasks can be found at Figure~\ref{fig:raven} and Figure~\ref{fig:mewl}. 

\begin{figure}[t]
    \centering
    \includegraphics[width=\linewidth]{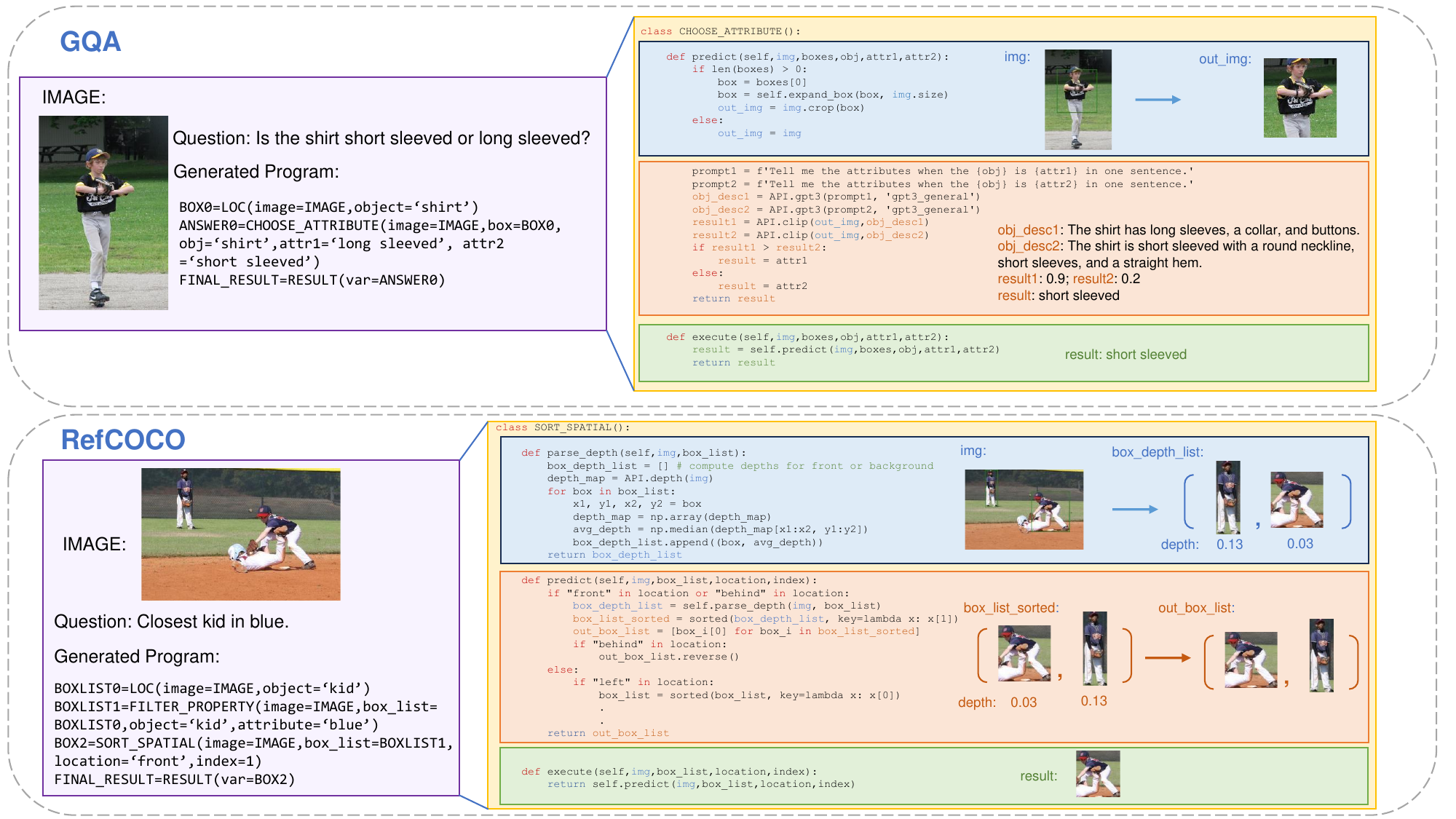}
    \caption{Qualitative examples of \Model's on GQA and RefCOCO. The query images, language instructions, and the parsed programs are shown on the left. The corresponding new modules and the value of important variables are shown on the right.}
    \label{fig:gqaRef}
\end{figure}

\subsection{Comparison with Baselines on Visual Reasoning.} 
We conducted analysis between our model and several baseline models using the GQA and RefCOCO datasets. Due to the deprecation of the original professional Codex API (\texttt{code-davinci-002}), we replaced it with the currently available API (\texttt{gpt-3.5-turbo-instruct}) and conducted experiments with both our model and the baseline models to ensure a fair comparison. We did not carry out experiments with GPT-4 due to the prohibitive cost. 
\input{tabs/compare}

The results, as presented in Table~\ref{tb:qa}, demonstrate that our model achieves competitive performance in both visual question answering and referring expression comprehension, thus confirming its effectiveness. Furthermore, we provide an illustrative module from our model in Figure~\ref{prompt:md1}. This newly created module has the capability to utilize various available APIs to select attributes from the images. The step-by-step reasoning process of our model is detailed in Figure~\ref{fig:gqaRef}, offering greater transparency compared to end-to-end models.

\begin{figure}[t]
    \centering
    \includegraphics[width=\linewidth]{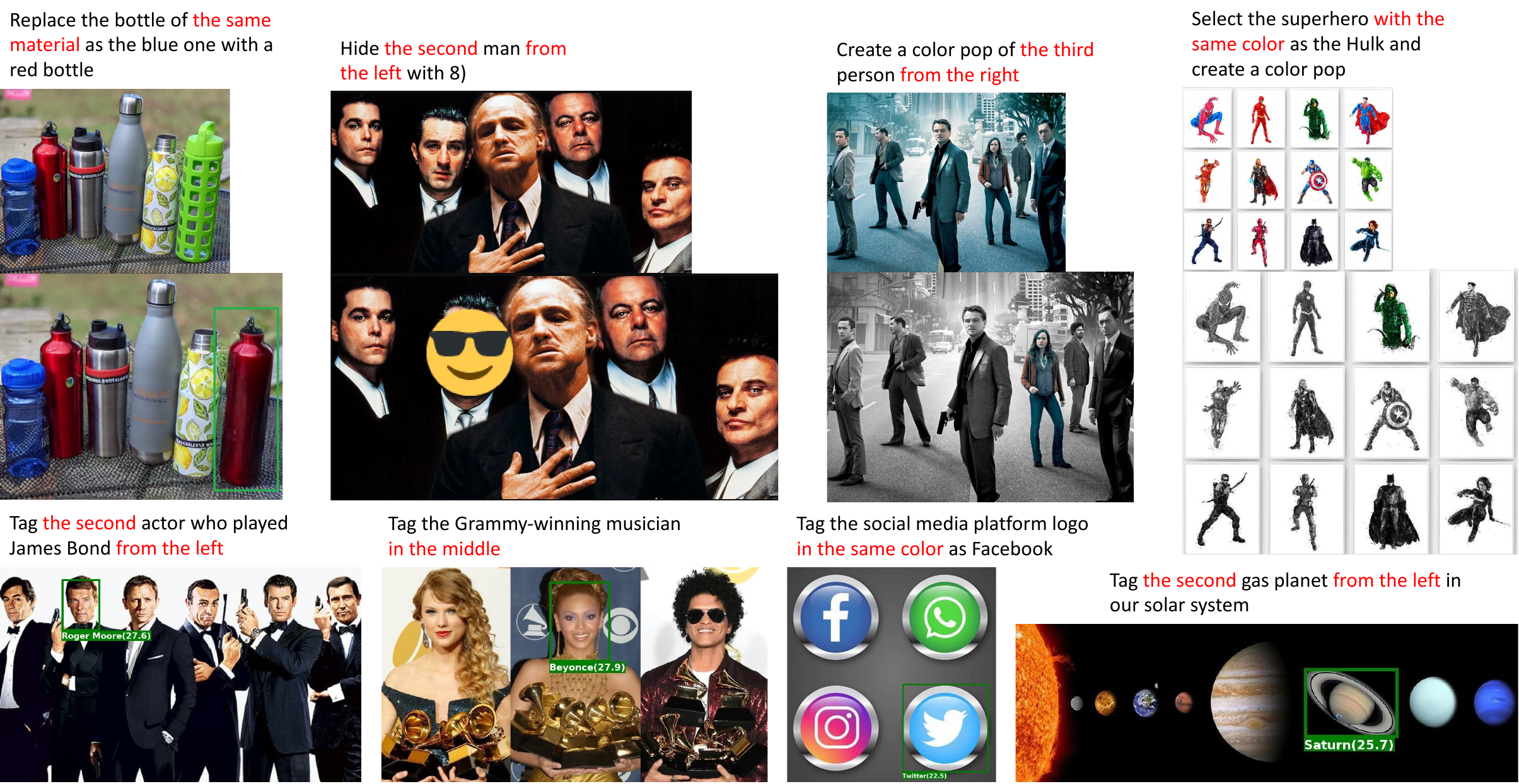}
    \vspace{-1em}
    \caption{Qualitative examples of \Model's on the image editing and knowledge tagging tasks. The language instructions of the tasks are shown the original images while the modified results of the images are shown below the original images. The emphasis of the instructions is highlighted with red colors, which requires our new modules to handle. While the key regions of the output images are bounded with green colors.}
    \label{fig:img_edit}
\end{figure}
\subsection{\Model for Transfer Learning}
In this section, we demonstrate our model's robust capabilities in transfer learning. We augment the modular library by incorporating modules created from GQA and RefCOCO, employing in-context examples to guide the Language Model (LLM) in generating step-by-step instructions for task execution. Qualitative results for this task are depicted in Figure~\ref{fig:img_edit}. As illustrated in Figure~\ref{fig:img_edit}, our model excels in generating semantically accurate images using the newly added module, whereas the baseline VisProg struggles to capture the required relationships with its fixed, pre-defined module library. To provide a more comprehensive evaluation of image editing, we enlist annotators to manually assess the correctness of the generated images. The models' performance is compared in Table~\ref{tb:trans}, where our model outperforms the baseline. In the context of knowledge tagging, we task annotators with marking image regions referenced by expressions and employ the same metrics as RefCOCO for evaluating the accuracy of bounding boxes and employ the BERT score to assess the correctness of labeled names. Our model demonstrates superior performance in both image editing and knowledge tagging. We show a typical example in Figure~\ref{contrast:tag} of the Appendix to show how our \Model make use of new modules to perform better knowledge tagging result the baseline.  

\input{tabs/transfer}
\subsection{\Model on Few-shot Task Learning.}
\begin{figure}[t]
\centering
\begin{minipage}{1.0\textwidth}
\centering
\includegraphics[width=\linewidth]{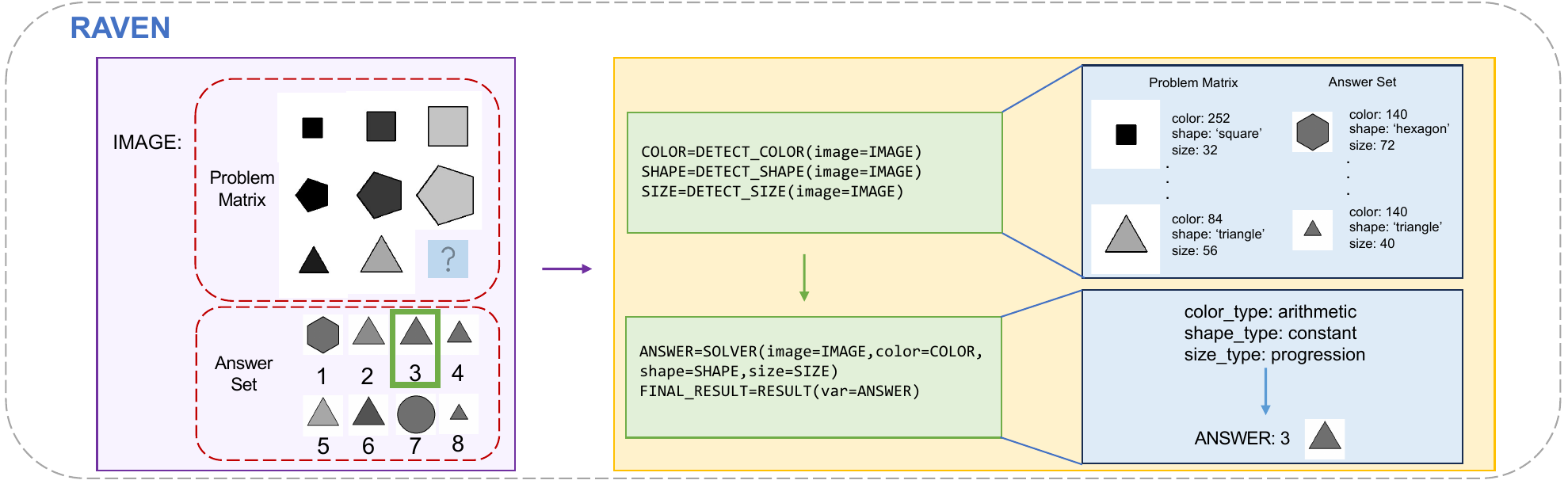}
\caption{
A qualitative example from the Raven dataset~\citep{zhang2019raven} is provided. This task involves a set of images with varying visual attributes, such as colors, shapes, and locations. Models are tasked with identifying the image that best matches the missing item in the Problem Matrix. \Model exhibits the capability to compose modules (\ie \texttt{DETECT\_SHAPE and SOLVER}) for detecting these attribute rules and constructing a solver module to address the task. The correct answer is indicated by a green box.
}
\label{fig:raven}
\end{minipage}
\begin{minipage}{1.0\textwidth}
\centering
\vspace{2em}
\includegraphics[width=\linewidth]{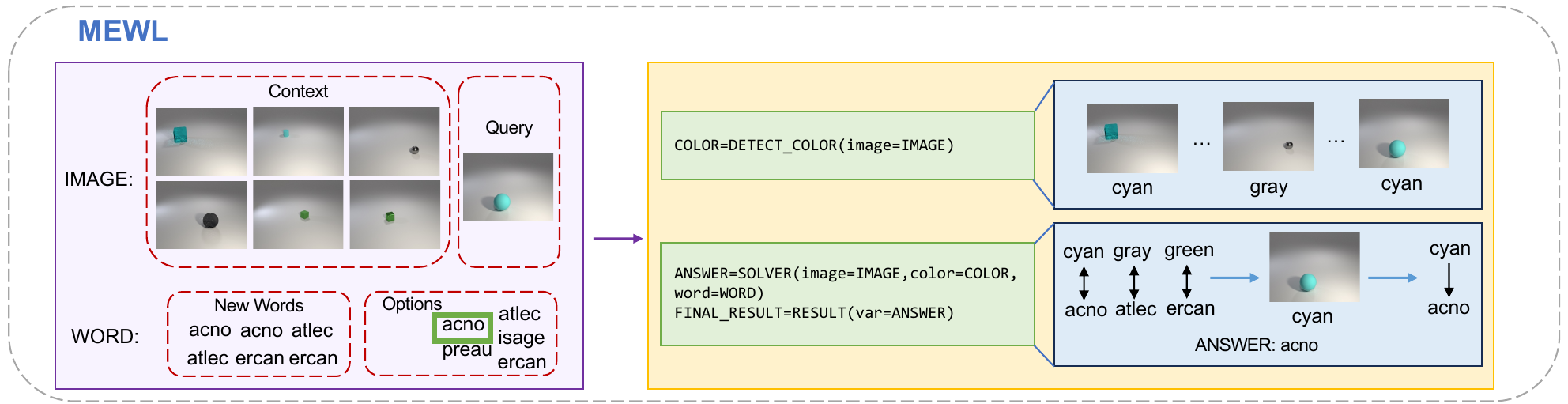}
\caption{
A qualitative illustration from the MEWL dataset~\citep{jiang2023mewl} is presented. This task entails a set of images featuring diverse visual attributes, such as material and shapes, and it necessitates models to determine the word that corresponds to the query image. \Model demonstrates the capability to generate modules for identifying these attribute rules and composing a solver module to address the task. The correct answer is indicated by a green box.
}
\label{fig:mewl}
\end{minipage}
\end{figure}

As a general module learning framework, our model is not only able to learn new modules to handle existing tasks but also can learn to handle new visual reasoning tasks from a few training examples.
We evaluate such abilities on new tasks, Raven~\citep{zhang2019raven} and MEWL~\citep{jiang2023mewl}. Specifically, we first prompt the LLM to learn pattern recognition modules for visual understanding and then ask the LLM to generate a solver module to handle the task.
The instances of our model prediction are shown in Figure~\ref{fig:raven} and Figure~\ref{fig:mewl}. Note that visual reasoning from Raven is widely used in intelligent testing for humans, which shows our model's strong capabilities and potential. We report the performance of our model and the baselines in Table~\ref{tb:raven} and Table~\ref{tb:mewl}. Our model is significantly better than previous fully-supervised methods like ResNet+DRT~\citep{zhang2019raven} and Aloe~\citep{ding2021attention}, showing its effectiveness. Note that all these models ResNet+DST, ALANS-V~\citep{zhang2022learning}, Aloe~\citep{ding2021attention} and Flamingo~\citep{alayrac2022flamingo} are models fully-finetuned on in-domain data, while our \Model is a general few-shot framework to learn modules for problem-solving. Moreover, we can observe the new compositionality and module re-usage from Figure~\ref{contrast:raven} of the Appendix. Although the SOLVER module was originally learned from center-type problems, it can be naturally transferred to other types like left-right and up-down.

\input{tabs/tab_new_tasks}

\subsection{Ablations}
\label{subsec:abs}
\input{tabs/abs}
To gauge the efficacy of our model, we conducted a series of ablation studies addressing the following key inquiries: \textbf{Q1} How effective is module learning? \textbf{Q2} What impact does the quantity of training examples have on model performance? \textbf{Q3} How crucial is the LLM's capability for optimal performance? In our experiments, \textit{\Model w/o ML} represents a configuration without any new module learning but relies heavily on ViperGPT and VisProg-defined modules, directing the LLM to pinpoint a region matching the referring expression. On the other hand, \textit{\Model-WLM} replaces the \texttt{gpt-3.5-turbo-instruct} API with \texttt{WizardCoder-Python-34B-V1.0} from WizardLM~\citep{xu2023wizardlm}. The designations \textit{\Model (10)/ (50) / (100)} indicate models trained with 10, 50, and 100 examples, respectively. For resource constraints, we limited our experimentation to 800 RefCOCO samples.

Table~\ref{tb:abs} presents the outcomes, leading to these insights: module learning, given sufficient test instances, can bolster task performance (\textbf{Q1} addressed). A paucity of training examples, such as 10 for RefCOCO, might induce overfitting, but this diminishes with increased training data (50 examples), improving overall performance (\textbf{Q2} addressed). Finally, model performance appears intrinsically tied to the LLM's capacity, with superior LLMs delivering enhanced results (\textbf{Q3} addressed).

\section{Conclusion}
In this study, we introduce \Model, which is designed to tackle visual reasoning tasks when confronted with limited training data. This approach combines language models to parse natural language into executable operations and create specialized visual modules tailored to the given task. Our model exhibits competitive performance on conventional tasks, effortless transfer of acquired modules to novel tasks, and the capability to adapt to new tasks even with limited training data. Our \Model also proposes numerous avenues for future research. Firstly, it still necessitates task-specific prompts for each distinct reasoning task, and it would be intriguing to explore the use of a universal prompt for all tasks. Secondly, the framework can be extended to encompass a broader range of multi-modal reasoning tasks, incorporating diverse inputs such as audio, video, and tactile information.

\bibliography{iclr2024_conference}
\bibliographystyle{iclr2024_conference}

\appendix
\newpage
\section{Appendix}
In this section, we substantiate our claims in the paper by providing additional implementation details (Section~\ref{imp:detail}), exemplar prompts for each stage (Section~\ref{prompt}), details on dataset collection (Section~\ref{dataset}), qualitative examples of new learned modules (Section~\ref{qualitative}).

\subsection{Implemenation Details}
\label{imp:detail}
\paragraph{Pre-defined Modules and API models.} The success of our model still requires a set of pre-defined APIs. Following the modules in VisProg and ViperGPT, we adopt the following APIs. We adopt GLIP~\citep{li2021grounded} for object localization. We adopt \textit{gpt-3.5-turbo-instruct} from OpenAI and  \textit{WizardCoder-Python-34B-V1.0} from WizardLM for code generation. We use BLIP~\citep{li2023blip} for answering simple questions about images. We use CLIP~\citep{Radford2021LearningTV} and X-VLM~\citep{xvlm} for image-text classification. We use MiDaS~\citep{Ranftl2021} for estimating depths in images. We use stable diffusion~\citep{rombach2022high} for modifying image patches. Based on these APIs, we construct a set of pre-defined modules following VisProg. These pre-defined modules will be used to cooperate with the new learned modules for visual reasoning.
\input{tabs/md}
\subsection{Prompts for each stage.}
\label{prompt}
The ability of Our \Model is from in-context learning of LLMs~\citep{brown2020language}, when the prompts are keys to tell what the LLM should generate. We show the exemplar prompts of our models to learn the VQA task in Figure~\ref{prompt:stage1}-\ref{prompt:stage3}.

\subsection{Details and Examples of the New Datasets.}
\label{dataset}
\begin{figure}[t]
    \centering
    \includegraphics[width=\linewidth]{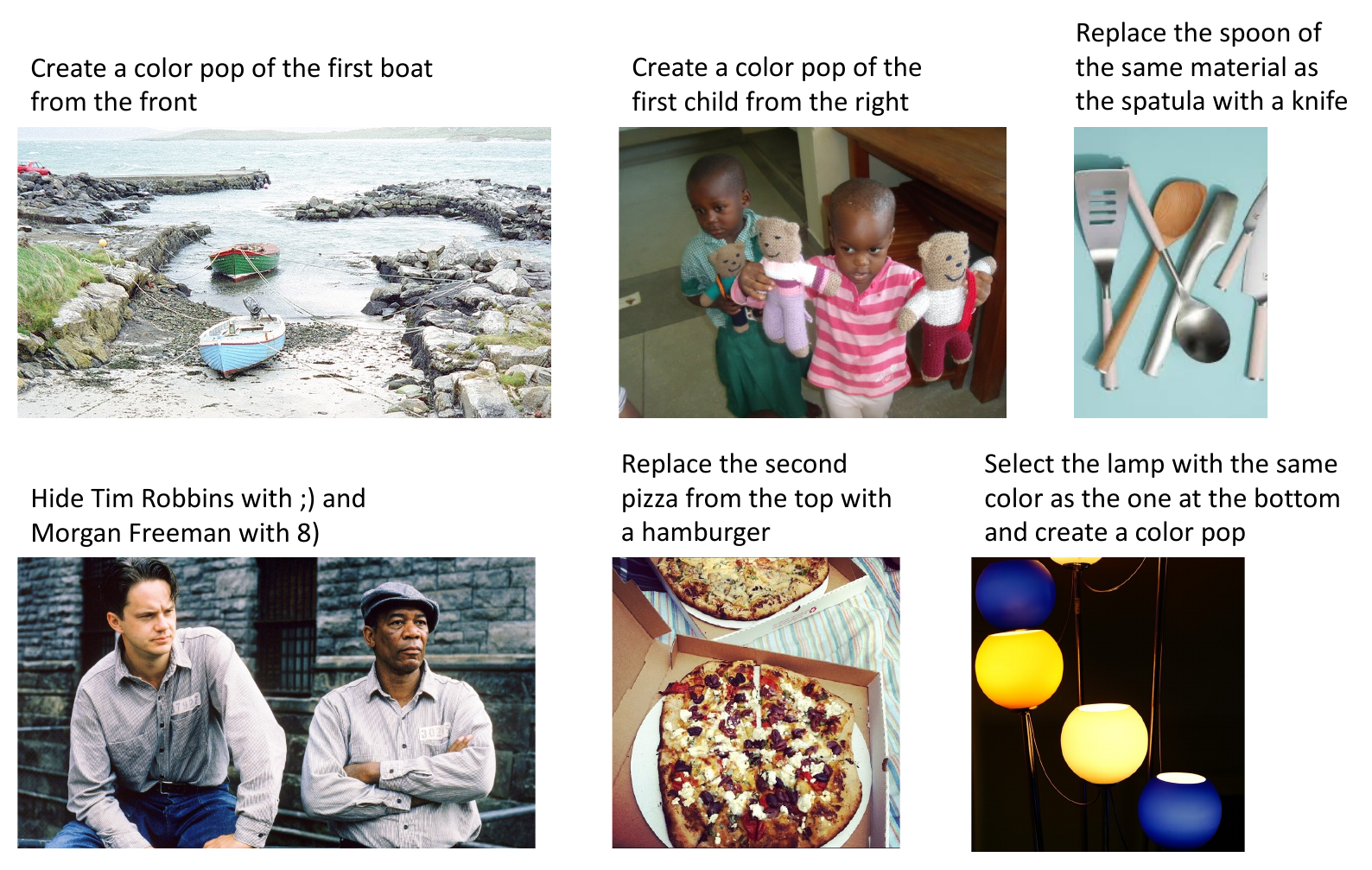}
    \caption{More examples of the new image edit dataset. The dataset asks models to edit images' fine-grained and regional details according to diverse language instructions.}
    \label{data:edit}
\end{figure}

\begin{figure}[t]
    \centering
    \includegraphics[width=\linewidth]{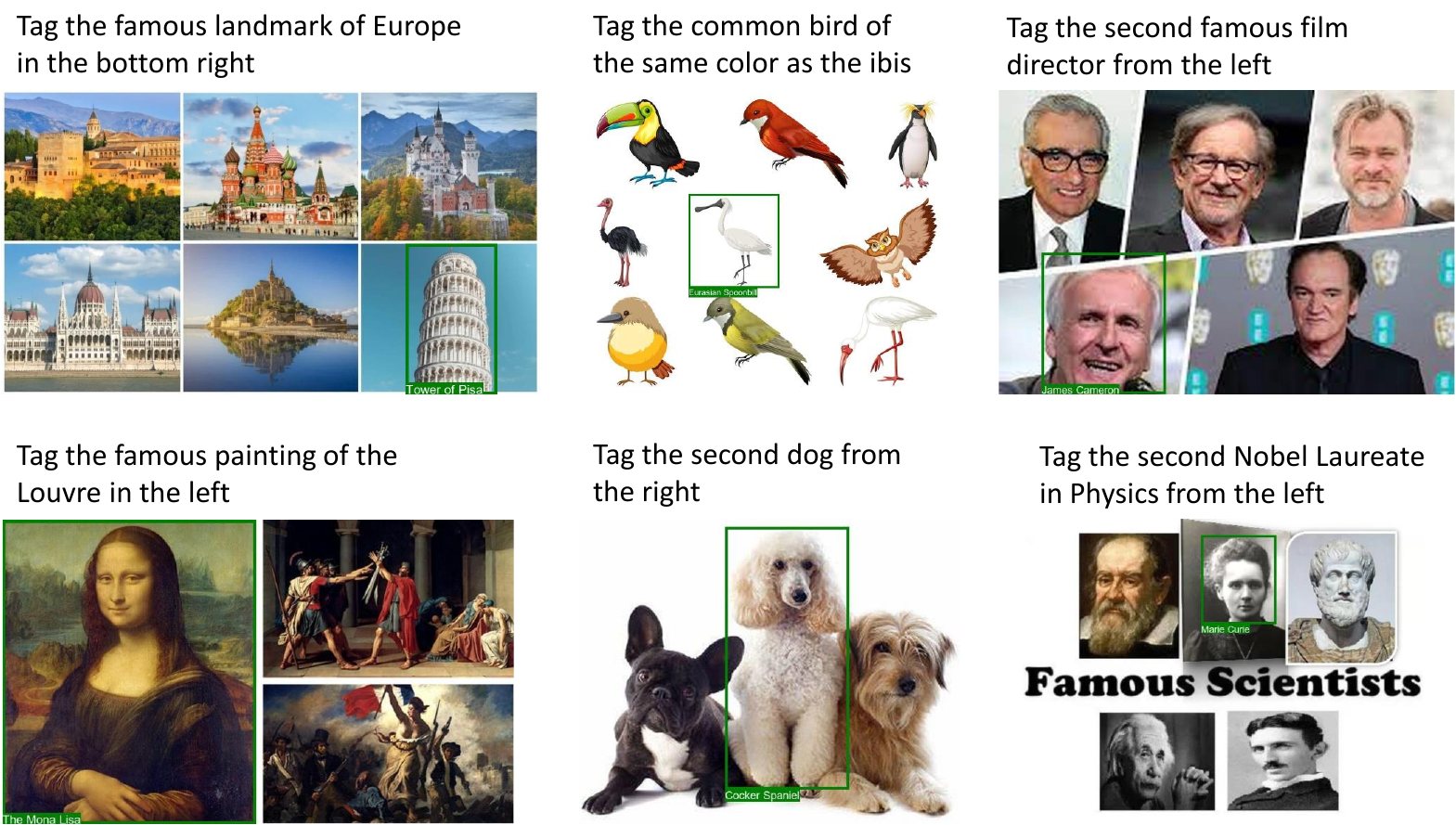}
    \caption{More examples of the new knowledge tagging dataset. The dataset requires models to localize the target region and tag the region with the desired information.} 
    \label{data:tag}
\end{figure}

To evaluate Knowledge Tagging, 50 tagging instructions are annotated on 50 internet images including personalities and a variety of objects such as logos, flowers, buildings, fruits and sports, among others. For each instruction, we manually annotated the ground truth bounding box and the associated tag.
For image editing assessment, we collected 50 editing instructions on 50 images including personalities and various objects like foods, furniture, animals, utensils \etc. 25 images are from the COCO dataset and the other 25 images are from the internet. For the image editing tasks, we ask three annotators to estimate whether the editing is correct or not. For the knowledge tagging task, we consider the localization is correct if the detected region has an IoU higher 0.5 with the ground-truth annotation. For text tagging, we compare the prediction with the annotated text with BERT matching (BEM)~\citep{bulian2022tomayto}. If the matching score is higher than 0.5, we consider it a successful matching.
More examples of the two datasets can be found at Figure~\ref{data:edit} and Figure~\ref{data:tag}.

\subsection{Qualitative Examples.}
\label{qualitative}
\begin{figure}[t]
    \centering
    \includegraphics[width=\linewidth]{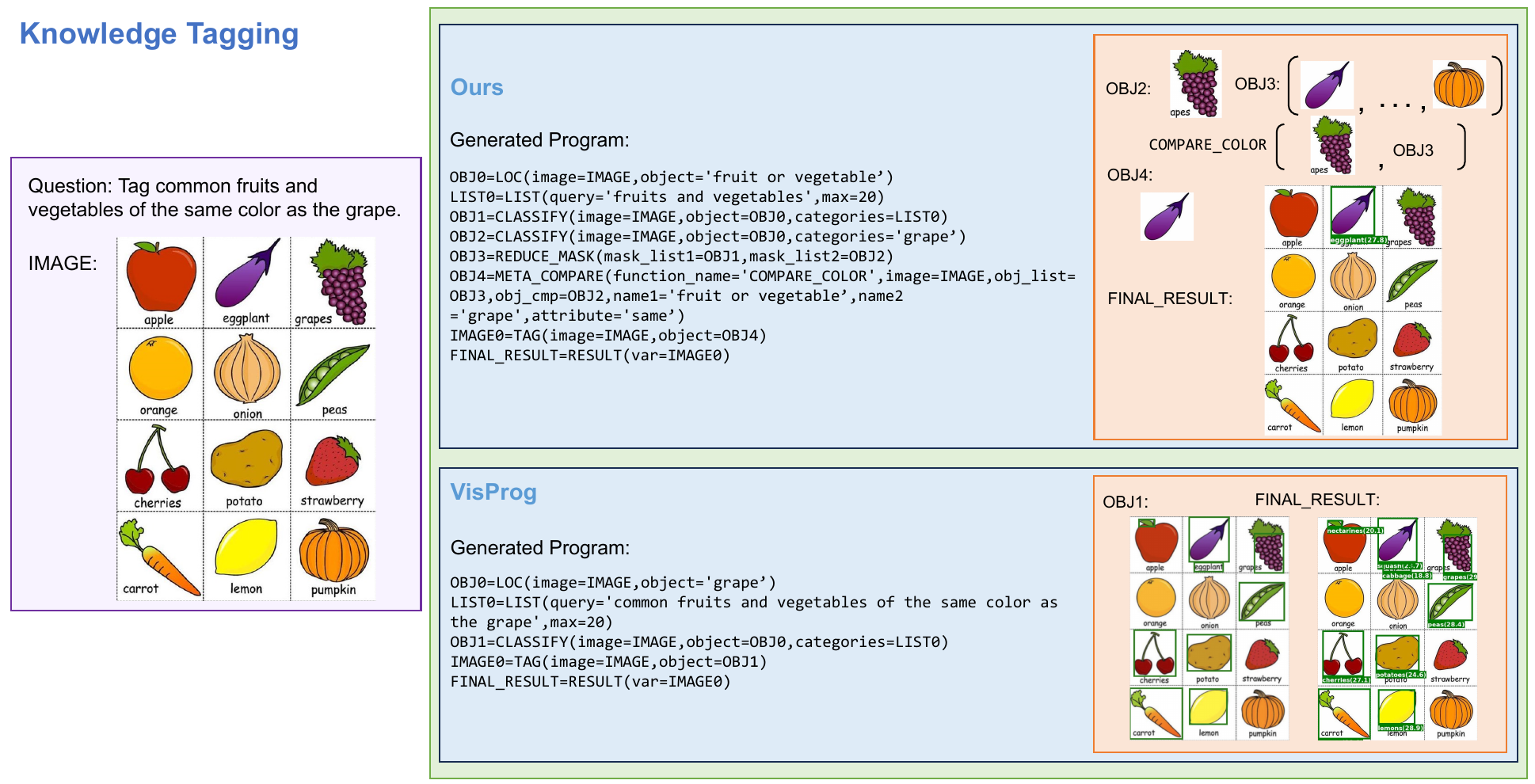}
    \caption{A typical example of how our \Model outperforms VisProg on knowledge tagging. In the top, our model is able to make use of the \textit{COMPARE\_COLOR} module learned from GQA to localize the correct region while VisProg fail to generate the correct program with its fixed module library.}
    \label{contrast:tag}
\end{figure}

\begin{figure}[t]
    \centering
    \includegraphics[width=\linewidth]{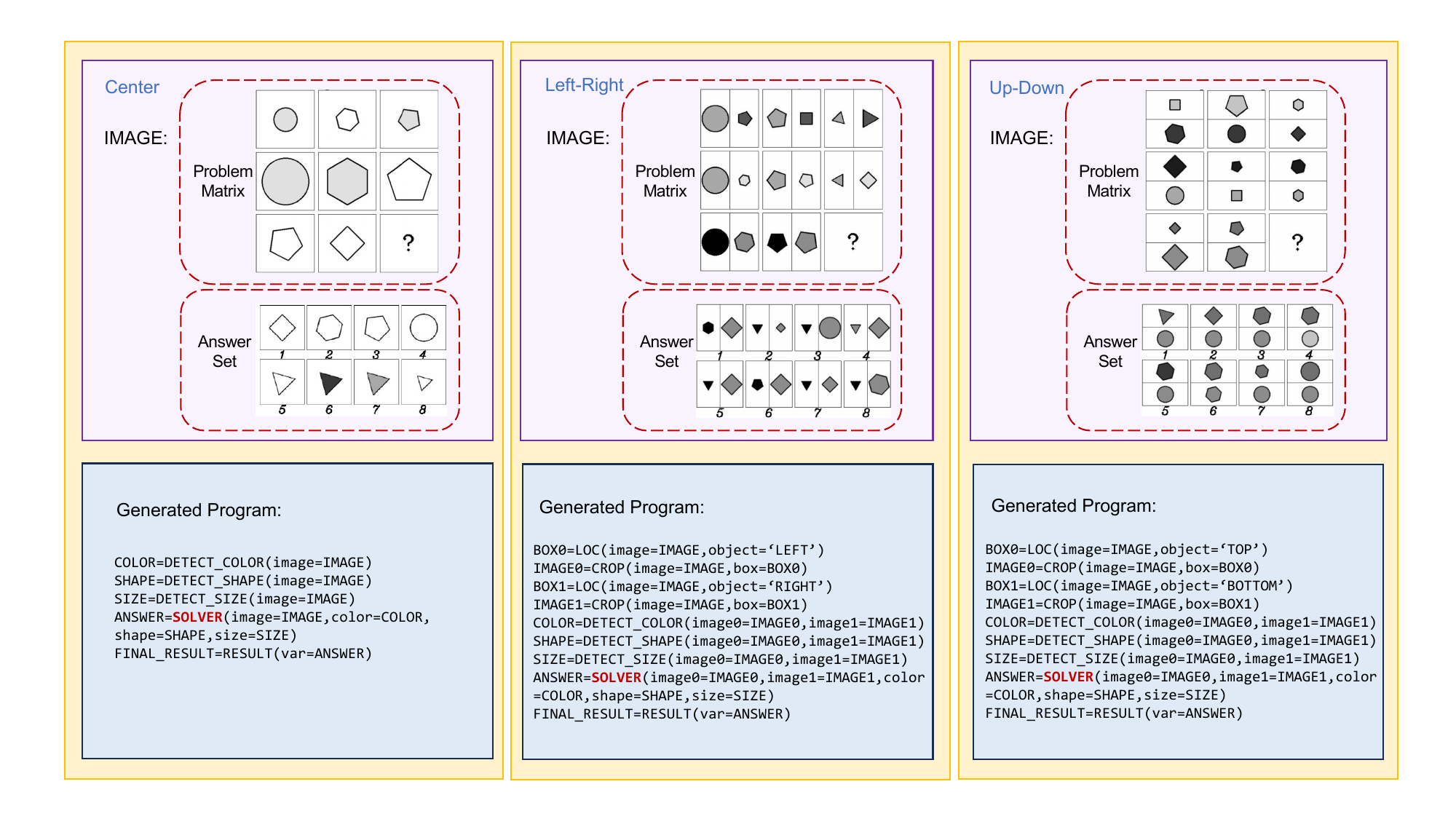}
    \caption{New compositionality and module re-usage in the Raven dataset. While the SOLVER module was initially trained on center-type problems in the Raven dataset, it exhibits a natural transferability to other types, such as left-right and up-down problems.}
    \label{contrast:raven}
\end{figure}

In this subsection, we show the qualitative examples of the learned modules and qualitative cases of how they handle different tasks.
We show an example of \Model performs better than VisProg in Figure~\ref{contrast:tag}. At the top of Figure~\ref{contrast:tag}, our model effectively utilizes the COMPARE\_COLOR module acquired from GQA to pinpoint the correct region, whereas VisProg fails to generate the correct program due to its rigid module library. Figure~\ref{contrast:raven} highlights emerging forms of compositionality and module re-usage. Notably, although the \textit{SOLVER} module was originally trained on center-type problems within the Raven dataset, it demonstrates inherent adaptability to other problem types, including left-right and up-down orientations.

\paragraph{New Learned Modules.} We show the examplar new learned modules from the GQA and RefCOCO in Figure~\ref{prompt:md1}-\ref{fig:sort2}. As shown in Figure~\ref{prompt:md1}, the new learned module ( \textcolor{blue}{CHOOSE\_ATTRIBUTE}) is able to use the LLM to retrieve relevant knowledge first and then adopt the image-text classifier to match the attributes. In Figure~\ref{fig:sort}-\ref{fig:sort2}, we see that the new module \textcolor{blue}{SORT\_SPATIAL} is able to localize objects with spatial index.

\input{modules/gqa_md1}
\input{modules/gqa_md2}
\input{modules/sort_module}
\input{modules/sort_module2}

\input{prompts/vqa_stage1}

\input{prompts/vqa_stage2}

\input{prompts/vqa_stage3}

\end{document}

%% file: tabs/compare.tex
\begin{wraptable}{r}{6cm}
\begin{tabular}{lcc}
\toprule
Methods & GQA & RefCOCO  \\
\midrule
BLIP-2  & 44.7 & -\\
KOSMOS-2 & - & 57.4 \\
ViperGPT-CodeX & 48.1 & 72.0 \\
\midrule
VisPROG-Instruct  & 45.4 & - \\
ViperGPT-Instruct & 38.2 & 62.4 \\
Ours-Instruct & 45.6 & 69.2 \\
\bottomrule
\end{tabular}
\vspace{-1em}
\caption{Evaluation on standard visual reasoning benchmarks, GQA and RefCOCO.}
\label{tb:qa}
\end{wraptable}

%% file: tabs/transfer.tex
\begin{table}[t]
\centering
\begin{tabular}{lcccccccc}
\toprule
\multirow{2}{*}{Methods} & \multicolumn{1}{c}{Image Editing} & \multicolumn{3}{c}{Tagging} & \multicolumn{3}{c}{Localization} \\
                         & Accuracy                          & Precision     & Recall     & F1       & Precision    & Recall   & F1     \\ 
\midrule
VisProg                  & 16.7                              & 18.4          & 21.7       & 19.9     & 32.8         & 35.3     & 34.0   \\
\Model                   & 55.3                              & 67.1          & 52.3       & 58.8     & 76.9         & 57.9     & 66.0   \\
\bottomrule
\end{tabular}
\caption{Evaluation of \Model on Transfer Learning with image editing and knowledge tagging tasks. Our \Model shows much better performance on all criteria, showing the effectiveness of the transferred modules. A qualitative comparison can be seen Figure~\ref{contrast:tag} in the Appendix.}
\label{tb:trans}
\end{table}

%% file: tabs/tab_new_tasks.tex
\begin{table}[t]
\centering
\parbox{.49\linewidth}{
\setlength{\tabcolsep}{7pt} 
\begin{tabular}{lcccccccc}
\toprule
Methods & \texttt{Center} & \texttt{L-R} & \texttt{U-D}  \\
\midrule
ResNet+DRT & 58.1 & 65.8 & 67.1 \\
ALANS-V & 98.4 & 97.3 & 96.4 \\
\midrule
\Model & 80.1 & 67.6 & 69.1\\
\midrule
Human & 95.5 & 86.4 & 81.8 \\
\bottomrule
\end{tabular}
\caption{Evaluation of \Model on Raven~\citep{zhang2019raven}. Compared with methods trained with massive in-domain data, our model performs competitively.}
\label{tb:raven}
}
\label{tb:block_qa}
\hfill
\parbox{.49\linewidth}{
\setlength{\tabcolsep}{3pt}
\begin{tabular}{lcccccccc}
\toprule
Methods & \texttt{shape} & \texttt{color} & \texttt{material}  \\
\midrule
Aloe & 34.2 & 33.2 & 31.0 \\
Flamingo-1.1B & 49.3 & 35.3 & 48.5 \\
\midrule
\Model & 43.7 & 45.3 & 41.0\\
\midrule
Human & 92.4 & 87.2 & 72.7 \\
\bottomrule
\end{tabular}
\caption{Evaluation of \Model on MEWL~\citep{jiang2023mewl}. Compared with approaches trained on extensive in-domain data, our model shows competitive performance.
}
\label{tb:mewl}
}
\vspace{-1em}
\end{table}

%% file: tabs/abs.tex
\begin{wraptable}{r}{6cm}
\vspace{-3em}
\begin{center}
\small
\begin{tabular}{lc}
\toprule
Methods  &  RefCOCO \\
\midrule
\Model w/o ML &   62.3  \\
\Model-WLM &  64.4  \\
\Model (10) & 49.4 \\
\Model (50) & 67.0 \\
\Model (100) & 67.1 \\
\bottomrule
\end{tabular}
\end{center}
\vspace{-1em}
\caption{Ablation study of \Model on RefCOCO.}
\vspace{-1em}
\label{tb:abs}
\end{wraptable}

%% file: tabs/md.tex
\begin{table}[h]
    \centering
    \begin{tabular}{l|l}
        \toprule
        Descriptions & Modules  \\
        \midrule
        \multirow{3}{*}{Image Understanding} & \textbf{Loc} for object location, \textbf{FaceDet} for face detection, \textbf{Select} and   \\
            & \textbf{Filter\_Property} for image-text classification. \textbf{Filter\_Spatial} for  \\
            & selecting image regions;  \\
        \multirow{4}{*}{Image Manipulation} & \textbf{Replace} for image editing, \textbf{colorPop} for changing images colors,   \\
            &  \textbf{BgBlur} for blurring background, \textbf{Tag} for annotating box regions \\
            &  and \textbf{Emoji} for face tagging. \textbf{Crop} and its variants for cropping \\
            &  patches from the images. \\
         \multirow{3}{*}{Others} & \textbf{List} for retrieving factual knowledge, \textbf{Count} for counting object \\
         & numbers, \textbf{Eval}, \textbf{Result}, \textbf{BOX2MASK} and \textbf{MASK2BOX} for \\
         & formatting outputs. \\
    \bottomrule
    \end{tabular}
    \caption{Pre-defined Modules used in \Model.}
    \label{tb:api}
\end{table}

%% file: modules/gqa_md1.tex
\begin{figure}
 \begin{minipage}[t]{\linewidth}
        \begin{minted}[
        frame=lines,
        framesep=2mm,
        fontsize=\footnotesize,
        bgcolor=Gray,
        numbersep=5pt,
        escapeinside=||,
        linenos
        ]{python}
class CHOOSE_ATTRIBUTE():
	"""
	Input:
        image: an image object
        box: a list of bounding boxes
        object: a string
        attribute1: a string
        attribute2: a string
    Output:
        result: a string
	Examples:
        Question: Is the coat thick or thin?
        BOX0=LOC(image=IMAGE,object='coat')
        ANSWER0=CHOOSE_ATTRIBUTE(image=IMAGE,box=BOX0,object='coat',
        attribute1='thick',attribute2='thin')
        FINAL_RESULT=RESULT(var=ANSWER0)
    """
    step_name = 'CHOOSE_ATTRIBUTE'

    def __init__(self):
        print(f'Registering {self.step_name} step')

    def expand_box(self,box,img_size,factor=1.5):
        W,H = img_size
        x1,y1,x2,y2 = box
        dw = int(factor*(x2-x1)/2)
        dh = int(factor*(y2-y1)/2)
        cx = int((x1 + x2) / 2)
        cy = int((y1 + y2) / 2)
        x1 = max(0,cx - dw)
        x2 = min(cx + dw,W)
        y1 = max(0,cy - dh)
        y2 = min(cy + dh,H)
        return [x1,y1,x2,y2]

    def predict(self,img,boxes,obj,attr1,attr2):
        if len(boxes) > 0:
            box = boxes[0]
            box = self.expand_box(box, img.size)
            out_img = img.crop(box)
        else:
            out_img = img
        prompt1 = f'Tell me the attributes when the {obj} is {attr1} in 
        one sentence.'
        prompt2 = f'Tell me the attributes when the {obj} is {attr2} in
        one sentence.'
        obj_desc1 = API.gpt3(prompt1, 'gpt3_general')
        obj_desc2 = API.gpt3(prompt2, 'gpt3_general')
        result1 = API.clip(out_img,obj_desc1)
        result2 = API.clip(out_img,obj_desc2)
        if result1 > result2:
            result = attr1
        else:
            result = attr2
        return result

    def execute(self,img,boxes,obj,attr1,attr2):
        result = self.predict(img,boxes,obj,attr1,attr2)
        return result
     \end{minted}
    \centering
\end{minipage}
    \caption{Exemplar generated module from the GQA dataset. This automatically constructed module can make use of different APIs to compare attributes of an image region.
    \label{prompt:md1}
    }
\end{figure}

%% file: modules/gqa_md2.tex
\begin{figure}
 \begin{minipage}[t]{1\linewidth}
        \begin{minted}[
        frame=lines,
        framesep=2mm,
        fontsize=\footnotesize,
        bgcolor=Gray,
        numbersep=5pt,
        escapeinside=||,
        linenos
        ]{python}
class COMPARE_COLOR():
    """
    Input:
        image: an image object
        box1: a list of bounding boxes
        box2: a list of bounding boxes
        object1: a string
        object2: a string
        compare_type: a string
    Output:
        result: a string
	"""
    def expand_box(self,box,img_size,factor=1.5):
        W,H = img_size
        x1,y1,x2,y2 = box
        dw = int(factor*(x2-x1)/2)
        dh = int(factor*(y2-y1)/2)
        cx = int((x1 + x2) / 2)
        cy = int((y1 + y2) / 2)
        x1 = max(0,cx - dw)
        x2 = min(cx + dw,W)
        y1 = max(0,cy - dh)
        y2 = min(cy + dh,H)
        return [x1,y1,x2,y2]
    def predict(self,img,boxes1,boxes2,obj1,obj2,compare_type):
        if len(boxes1) > 0:
            box1 = boxes1[0]
            box1 = self.expand_box(box1,img.size)
            out_img1 = img.crop(box1)
        else:
            out_img1 = img
        if len(boxes2) > 0:
            box2 = boxes2[0]
            box2 = self.expand_box(box2,img.size)
            out_img2 = img.crop(box2)
        else:
            out_img2 = img
        color1 = API.vqa(out_img1, f'What color is the {obj1}?')
        color2 = API.vqa(out_img2, f'What color is the {obj2}?')
        prompt = f'Can the {color1} be regarded as the same color as' 
        f'{color2}? You should just reply yes or no without any other
        words.'
        temp = API.gpt3(prompt, 'gpt3_general')
        if 'same' == compare_type:
            if 'yes' in temp.lower():
                result = 'yes'
            elif 'no' in temp.lower():
                result = 'no'
        elif 'different' == compare_type:
            if 'yes' in temp.lower():
                result = 'no'
            elif 'no' in temp.lower():
                result = 'yes'
        else:
            if 'yes' in temp.lower():
                result = 'yes'
            elif 'no' in temp.lower():
                result = 'no'
        return result
    def execute(self,img,boxes1,boxes2,obj1,obj2,compare_type):
        result = self.predict(img,boxes1,boxes2,obj1,obj2,compare_type)
        return result 
     \end{minted}
    \centering
\end{minipage}
    \vspace{-1.5em}
     \caption{Exemplar generated module from the GQA dataset.
    \label{prompt:md2}
    }
\end{figure}

%% file: modules/sort_module.tex
\begin{figure}
 \begin{minipage}[t]{\linewidth}
        \begin{minted}[
        frame=lines,
        framesep=2mm,
        fontsize=\footnotesize,
        bgcolor=Gray,
        numbersep=5pt,
        escapeinside=||,
        linenos
        ]{python}
class SORT_SPATIAL():
    """
    Select objects from the image that match the spatial location.
    Objects are represented by the bounding boxes.
    Returns the bounding boxes that satisfie the condition.
    Input:
        image: raw PIL image
        box_list: a list of unormalized bounding boxes
        location: the location can only be left, middle, right, top,
        bottom, front and behind
        index: a number for the rank the object
    Output:
        box: a bounding box
    Examples:
        Question: second sandwich from the right on the bottom
        BOXLIST0=LOC(image=IMAGE,object='sandwich')
        BOXLIST1=SORT_SPATIAL(image=IMAGE,box_list=BOXLIST0,location=
        'right',index=2)
        BOXLIST2=SORT_SPATIAL(image=IMAGE,box_list=BOXLIST1,location=
        'bottom',index=1)
        FINAL_RESULT=RESULT(var=BOXLIST2)
    """
    step_name = 'SORT_SPATIAL'
    def predict(self,img,box_list,location,index):
        if index < 0 or index > len(box_list):
            return []
        if index == 0:
            return [box_list[0]]
        if "front" in location or "behind" in location:
            box_depth_list = self.parse_depth(img, box_list)
            box_list_sorted = sorted(box_depth_list, key=lambda x: x[1])
            out_box_list = [box_i[0] for box_i in box_list_sorted]
            if "behind" in location:
                out_box_list.reverse()
        else:
            if "left" in location:
                box_list = sorted(box_list, key=lambda x: x[0])
            elif "right" in location:
                box_list = sorted(box_list, key=lambda x: x[2], reverse
                    =True)
            elif "top" in location:
                box_list = sorted(box_list, key=lambda x: x[1])
            elif "bottom" in location:
                box_list = sorted(box_list, key=lambda x: x[3], reverse
                    =True)
            else:
                return []
            if index > len(box_list):
                return []
            out_box_list = [box_list[index-1]]
        return out_box_list
    def check_location(self,img,box,location):
        w, h = img.size
        x1, y1, x2, y2 = box
        cx = (x1 + x2) / 2
        cy = (y1 + y2) / 2
        if 'left' in location:
            if cx > w / 2:
                return False

     \end{minted}
    \centering
\end{minipage}
    \caption{Exemplar generated module from the RefCOCO dataset. The rest part of the code is in Figure~\ref{fig:sort2}.
    \label{fig:sort}
    }
\end{figure}

%% file: modules/sort_module2.tex
\begin{figure}
 \begin{minipage}[t]{\linewidth}
        \begin{minted}[
        frame=lines,
        framesep=2mm,
        fontsize=\footnotesize,
        bgcolor=Gray,
        numbersep=5pt,
        escapeinside=||,
        linenos
        ]{python}
        elif 'right' in location:
            if cx < w / 2:
                return False
        if 'top' in location:
            if cy > h / 2:
                return False
        elif 'bottom' in location:
            if cy < h / 2:
                return False
        return True

    def parse_depth(self,img,box_list):
        box_depth_list = [] 
        # compute depths for front or background
        depth_map = API.depth(img)
        for box in box_list:
            x1, y1, x2, y2 = box
            depth_map = np.array(depth_map)
            avg_depth = np.median(depth_map[x1:x2, y1:y2])
            box_depth_list.append((box, avg_depth))
        return box_depth_list

    def execute(self,img,box_list,location,index):
        return self.predict(img,box_list,location,index)
     \end{minted}
    \centering
\end{minipage}
    \caption{Exemplar generated module from the RefCOCO dataset. The former part of the code is in Figure~\ref{fig:sort}. This generated module is able to localize objects based on their location in images and the depth of images.
    }
    \label{fig:sort2}
\end{figure}

%% file: prompts/vqa_stage1.tex
\begin{figure}
 \begin{minipage}[t]{\linewidth}
        \begin{minted}[
        frame=lines,
        framesep=2mm,
        fontsize=\footnotesize,
        bgcolor=Gray,
        numbersep=5pt,
        escapeinside=||,
        linenos
        ]{python}

Pre-defined Modules:
class LOC():
    """
    Generate boxes of the object on the image.
    Input:
        image: an image object
        object: an object string
    Output:
        box: a list of bounding boxes
	Examples:
	   BOX0=LOC(image=IMAGE,object='camel')
	"""
class COUNT():
    """
    Count the number of boxes in the list.
    Input:
        box: a list of bounding boxes
    Output:
        number: number of boxes
    Examples:
	ANSWER0=COUNT(box=BOX1)
    """
Suppose you are a program expert. Given a set of pre-defined modules, 
could you identify whether it is possible to write a program to get the
answer to the question? If not, what new modules do we need?
Note that you can only use the below pre-defined modules:
LOC, COUNT, CROP .......

Question: Is the purse to the left or to the right of the person?
Yes. The program is:
BOX0=LOC(image=IMAGE,object='person')
IMAGE0=CROP_LEFTOF(image=IMAGE,box=BOX0)
BOX1=LOC(image=IMAGE0,object='purse')
ANSWER0=COUNT(box=BOX1)
ANSWER1=EVAL(expr=f"'left' if {ANSWER0} > 0 else 'right'")
FINAL_RESULT=RESULT(var=ANSWER1)

Question: Which object is larger, the sphere or the blue cube?
No. We need to make a new module "COMPARE_SIZE" first. Here is the header
of the class:
class COMPARE_SIZE():
    """
    Compare the size of two objects in the image.
    One object is identified by the first bounding box of box0
    Another object is identified by the first bounding box of box1
    Input:
	image: an image object
        box0: a list of bounding boxes
        box1: a list of bounding boxes
    Output:
        flag: return True if first object is larger else False
    Examples:
	Question: Which object is larger, the sphere or the blue cube?
	BOX0=LOC(image=IMAGE,object='sphere')
	BOX1=LOC(image=IMAGE,object='blue cube')
	FLAG0=COMPARE_SIZE(image=IMAGE,box0=BOX0,box1=BOX1)
	ANSWER2=EVAL(expr=f"'sphere' if {FLAG0} else 'blue cube'")
	FINAL_RESULT=RESULT(var=ANSWER)
    """
.......
Question: __INSERT_NEW_QUESTION__
    \end{minted}
    \centering
\end{minipage}
    \vspace{-1.5em}
    \caption{Prompt of \ModuleA (Stage 1) to propose the module candidates.
    \label{prompt:stage1}
    }
\end{figure}

%% file: prompts/vqa_stage2.tex
\begin{figure}
 \begin{minipage}[t]{\linewidth}
        \begin{minted}[
        frame=lines,
        framesep=2mm,
        fontsize=\footnotesize,
        bgcolor=Gray,
        numbersep=5pt,
        escapeinside=||,
        linenos
        ]{python}
Pre-defined APIs:
class API():
    def loc(cls, image: Image.Image, obj_name: str) -> List[List[int]]:
        """
        Returns boxes of the specific object on the image.
        Input:
            image: an image object
            obj_name: a text string
        Output:
            boxes: a list of bounding boxes
        Example:
            [[261, 160, 525, 299]] = API.loc(image, "camel")
        """
        return boxes
    def gpt3(cls, prompt: str) -> str:
        """
        Returns response from gpt3
        Input:
            prompt: prompt to gpt3
        Output:
            response: gpt3 response
        Example:
            "joe biden" = API.gpt3("Who is the US President?")
        """
        return response
    ....... 

Given the above APIs, could you write a new module, Loc()?
class Loc():
    """
    Generate boxes of the object on the image.
    Input:
        img: an image object
        obj_name: an object string
    Output:
        selected_boxes: a list of bounding boxes
    Examples:
        BOX1=LOC(image=IMAGE0,object='food')
    """
Response:
class Loc():
    step_name = 'LOC'
    def execute(self,img,obj_name):
        if obj_name=='TOP':
            bboxes = [self.top_box(img)]
        elif obj_name=='BOTTOM':
            bboxes = [self.bottom_box(img)]
        elif obj_name=='LEFT':
            bboxes = [self.left_box(img)]
        elif obj_name=='RIGHT':
            bboxes = [self.right_box(img)]
        else:
            bboxes = self.predict(img,obj_name)
        return bboxes

    def predict(self,img,obj_name):
        return API.loc(img,obj_name)
    ....... 
Given the above APIs, could you write a new module, __MODULE_NAME__?
__MODULE_HEAD__   
    \end{minted}
    \centering
\end{minipage}
    \vspace{-2em}
    \caption{Prompt of \ModuleB (Stage 2) to make a module based on the module's input and output. 
    \label{prompt:stage2}
    }
\end{figure}

%% file: prompts/vqa_stage3.tex
\begin{figure}
 \begin{minipage}[t]{\linewidth}
        \begin{minted}[
        frame=lines,
        framesep=2mm,
        fontsize=\footnotesize,
        bgcolor=Gray,
        numbersep=5pt,
        escapeinside=||,
        linenos
        ]{python}

Think step by step to answer the question.
  
You can only use modules below:
LOC
COUNT
EVAL
RESULT
VERIFY_ATTRIBUTE
VERIFY_COLOR
VERIFY_MATERIAL
.......

Question: Is the vehicle in the top of the image?
Program:
BOX0=LOC(image=IMAGE,object='TOP')
IMAGE0=CROP(image=IMAGE,box=BOX0)
BOX1=LOC(image=IMAGE0,object='vehicle')
ANSWER0=COUNT(box=BOX1)
ANSWER1=EVAL(expr=f"'yes' if {ANSWER0} > 0 else 'no'")
FINAL_RESULT=RESULT(var=ANSWER1)

Question: Who is carrying the umbrella?
Program:
BOX0=LOC(image=IMAGE,object='umbrella')
IMAGE0=CROP(image=IMAGE,box=BOX0)
ANSWER0=VQA(image=IMAGE0,question='Who is carrying the umbrella?')
FINAL_RESULT=RESULT(var=ANSWER0)

Question: Do the towel and the box have a different colors?
Program:
BOX0=LOC(image=IMAGE,object='towel')
BOX1=LOC(image=IMAGE,object='box')
ANSWER0=COMPARE_ATTRIBUTE(image=IMAGE,box1=BOX0,box2=BOX1,object1='towel'
,object2='box',attribute='color',question=QUESTION)
FINAL_RESULT=RESULT(var=ANSWER0)

Question: Is the knife made of ceramic?
Program:
BOX0=LOC(image=IMAGE,object='knife')
ANSWER0=VERIFY_MATERIAL(image=IMAGE,box=BOX0,material='ceramic',object=
'knife',question=QUESTION)
ANSWER1=EVAL(expr=f"'yes' if {ANSWER0} else 'no'")
FINAL_RESULT=RESULT(var=ANSWER1)

Question: Is the coat thick or thin?
Program:
BOX0=LOC(image=IMAGE,object='coat')
ANSWER0=CHOOSE_ATTRIBUTE(image=IMAGE,box=BOX0,object='coat',attribute1=
'thick',attribute2='thin')
FINAL_RESULT=RESULT(var=ANSWER0)
.......

Question: __INSERT_NEW_QUESTION__
Program:

    \end{minted}
    \centering
\end{minipage}
    \caption{Prompt of \ModuleC (Stage 3) to parse programs for a new test case.
    \label{prompt:stage3}
    }
\end{figure}